\documentclass[letterpaper, 10 pt, conference]{ieeeconf}
\IEEEoverridecommandlockouts                             
\usepackage{graphicx}
\usepackage{cite}
\usepackage[export]{adjustbox}
\usepackage{amsmath} 
\usepackage[dvipsnames]{xcolor}
\usepackage{todonotes}
\usepackage{soul}
\usepackage[margin=1in]{geometry} 
\usepackage[colorlinks,citecolor=blue,urlcolor=blue,bookmarks=true,hypertexnames=true]{hyperref} 
\usepackage[hypcap=true,font=footnotesize,labelfont=bf]{caption}
\usepackage{lipsum}
\usepackage{float}
\makeatletter
\def\BState{\State\hskip-\ALG@thistlm}
\makeatother 
\usepackage[dvipsnames]{xcolor}
\usepackage{subfiles}
\usepackage{subcaption}
\usepackage{algpseudocode}
\usepackage{booktabs}
\usepackage[linesnumbered,ruled,vlined]{algorithm2e} 
\usepackage{multirow}

\title{\LARGE \bf
Optimal Constrained Task Planning as Mixed Integer Programming
}

\author{Alphonsus Adu-Bredu$^{1}$ \hspace{0.5cm} Nikhil Devraj$^{1}$ \hspace{0.5cm}   Odest Chadwicke Jenkins$^{1}$%
\thanks{$^{1}$Alphonsus Adu-Bredu, Nikhil Devraj and Odest Chadwicke Jenkins are with the Robotics Institute and Department of Electrical Engineering and Computer Science, University of Michigan, Ann Arbor, MI, USA.
        {\tt\small [adubredu|devrajn|ocj]@umich.edu}}%
}

\begin{document}

\maketitle
\thispagestyle{empty}
\pagestyle{empty}

\begin{abstract}
For robots to successfully execute tasks assigned to them, they must be capable of planning the right sequence of actions. These actions must be both optimal with respect to a specified objective and satisfy whatever constraints exist in their world. We propose an approach for robot task planning that is capable of planning the optimal sequence of grounded actions to accomplish a task given a specific objective function while satisfying all specified numerical constraints. Our approach accomplishes this by encoding the entire task planning problem as a single mixed integer convex program, which it then solves using an off-the-shelf Mixed Integer Programming solver. We evaluate our approach on several mobile manipulation tasks in both simulation and on a physical humanoid robot. Our approach is able to consistently produce optimal plans while accounting for all specified numerical constraints in the mobile manipulation tasks. Open-source implementations of the components of our approach as well as videos of robots executing planned grounded actions in both simulation and the physical world can be found at this url: \href{https://adubredu.github.io/gtpmip}{https://adubredu.github.io/gtpmip}
\end{abstract}

\section{Introduction}
The successful execution of manual tasks often requires the satisfaction of certain physical constraints. For instance, to retrieve a sugar canister from a seven-foot high shelf in a kitchen, the average person would have to stand \textit{close enough} and stretch their hands \textit{far enough} to not only reach the sugar, but to grasp it stably and lift it. Here, \textit{close enough} and \textit{far enough} are physical constraints that need to be satisfied to guarantee the success of their efforts to retrieve the sugar. Similarly, for a robot to successfully perform this sugar retrieval task, it would have to account for similar physical constraints when \textit{deciding} the \textit{right actions} to take. While deciding the right actions, this shrewd robot would also have to bias its decisions towards actions that optimize certain quantities like energy consumed or distance travelled. We call this problem the \textit{Optimal Constrained Task Planning} problem.

\begin{figure}
    \centering
    \includegraphics[width=\linewidth]{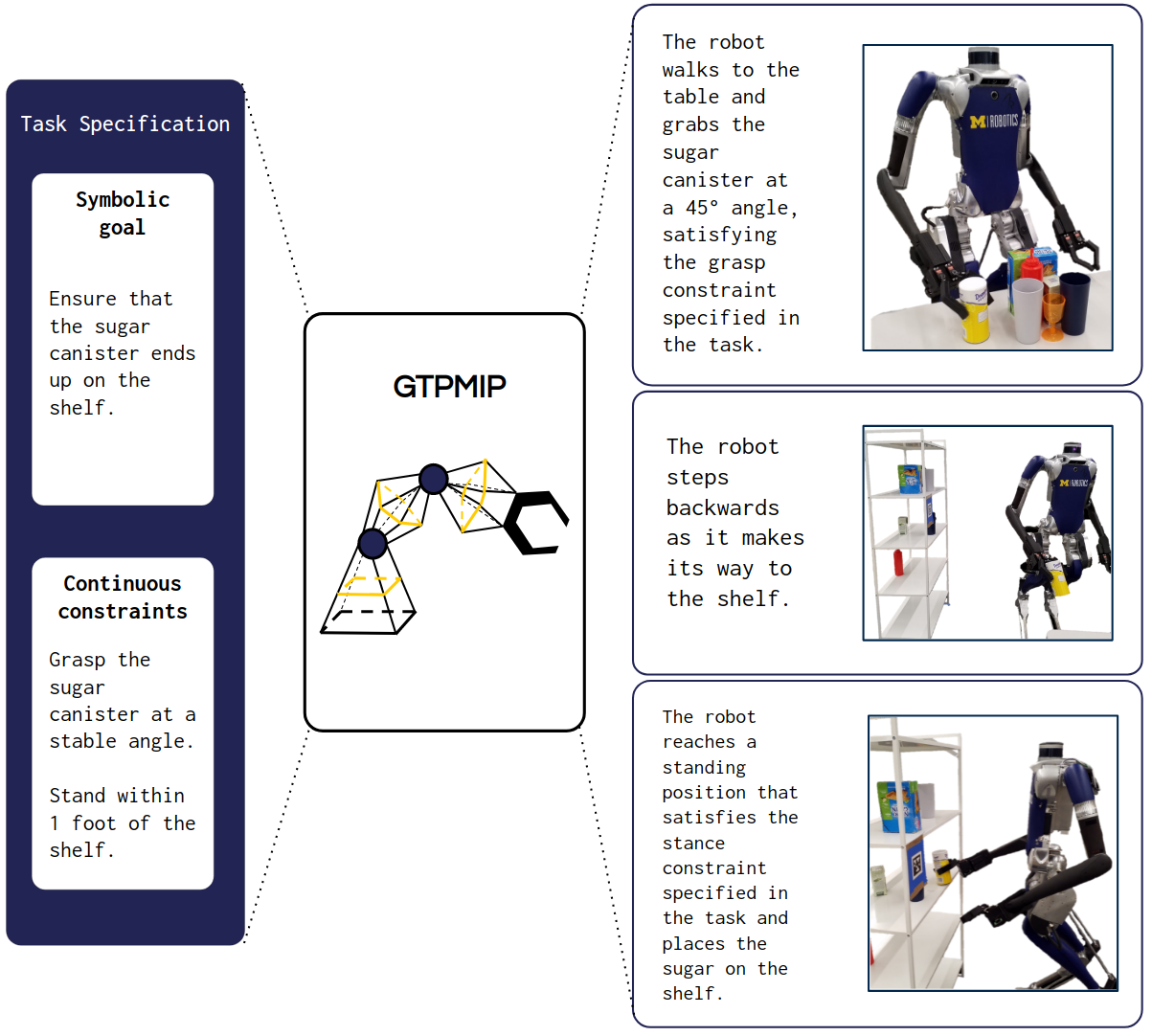}
    \caption{Given a constrained task planning problem, our approach (GTPMIP) plans a sequence of coherent actions with optimal parameters needed to accomplish the task.}
    \label{fig:teaser}
\end{figure} 
The predominant way to solve a task planning problem is to formulate it as a symbolic AI planning problem, represent it in a graph structure and employ graph search algorithms to find paths from a start state to a goal state. However, this approach is purely symbolic and provides no avenues for incorporating numerical constraints in the planning process or to bias the search to choose actions that optimize numerical objective functions \cite{pyperplan, fd, graphplan }. As such, these approaches only allow the specification of symbolic task goals and are unable to support the specification of continuous goals. It is often up to the human expert to introduce symbols that aptly represent desired continuous goals. A few works have proposed extensions to graph search algorithms to enable them to handle constraints and objective functions \cite{edelkamp2009optimal}. However, these approaches are only capable of handling simple additive objective functions with soft linear constraints.

In this work, we propose an approach for task planning that is capable of handling both linear and non-linear constraints and optimizes for convex objective functions to global optimality. We take a unique approach to the task planning problem by encoding the entire task planning problem as a single Mixed Integer Convex Program (MICP). By doing this, we gain the flexibility of subjecting the problem to arbitrary action constraints that need to be satisfied in order for the resulting plan to be physically realizable by a robot. We also escape the restriction of having to specify planning goals symbolically as this encoding enables the specification of continuous planning goals. We then use an off-the-shelf Mixed Integer Programming solver to solve the MICP to optimality, extract the grounded plan and its optimal parameters and execute the plan with a robot. The unique contributions of this work are as follows:
\begin{itemize}
    \item Firstly, we extend the Planning Domain Definition Language (PDDL) to allow for the specification of numerical action and task constraints, numerical initial values of continuous task variables, numerical objective functions, numerical action dynamics functions, numerical preconditions and numerical effects. We call this extension the Hybrid PDDL Description (HPD). 
    \item Secondly, we propose a unique representation of continuous actions as Funnels and propose an approach for representing the continuous plan space, which we call the Hybrid Funnel Graph.
    \item Finally, we describe an approach for encoding the Hybrid Funnel Graph as a single Mixed Integer Convex Program which we solve using an off-the-shelf MIP solver.
\end{itemize}

We evaluate our approach in simulation on several 2-D object rearrangement task planning problems subject to unique geometric constraints. We also demonstrate our approach on real-world mobile manipulation tasks involving kinematic constraints using the Digit humanoid robot, as shown in Figure \ref{fig:teaser}.

\begin{figure*}
    \centering
    \includegraphics[width=\textwidth]{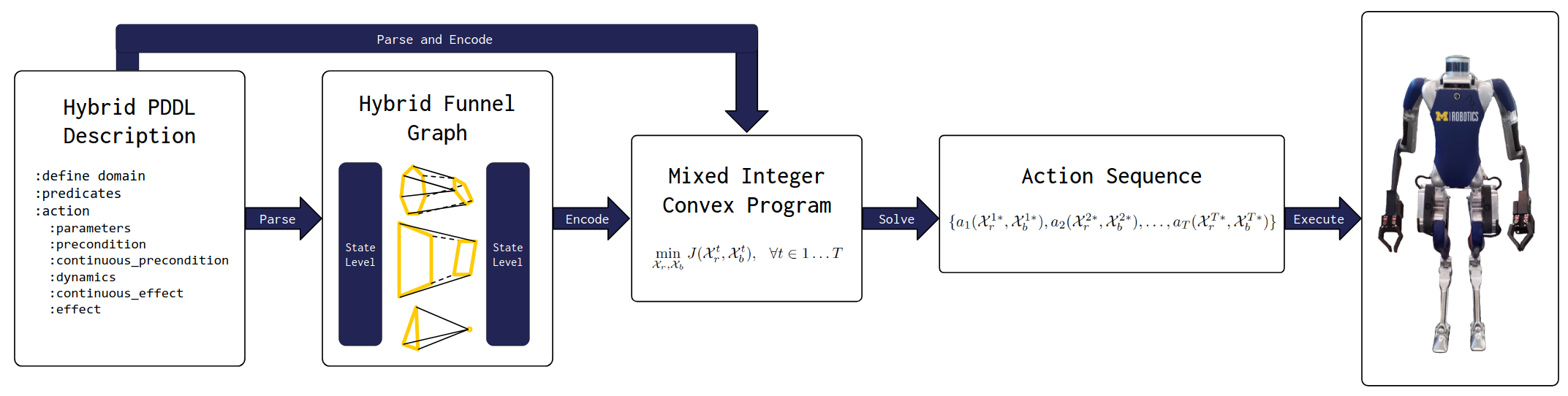}
    \caption{An overview of GTPMIP. Given a task, our approach represents the plan space as a Hybrid Funnel Graph, encodes the task and the Hybrid Funnel Graph as a Mixed Integer Convex Program and solves it to produce an optimal action sequence which is executed by the robot.}
    \label{fig:overview}
\end{figure*}
\section{Related Works}
\subsection{Mixed Integer Programming}
A Mixed Integer Program (MIP) is a mathematical optimization problem with both integer and real-valued variables \cite{lee2011mixed}. The ability of MIPs to have both discrete and continuous variables makes them ideal for formulating sequential planning problems that involve taking discrete actions which are subject to continuous constraints \cite{aceituno2020global, deits2014footstep, hogan2016feedback, valenzuela2016mixed}. This work encodes the optimal constrained task planning problem as a Mixed Integer Convex Program (MICP).

\subsection{Symbolic AI Planning}
The state-of-the-art methods in STRIPS-style \cite{strips} Symbolic AI Planning first decompose the planning problem into a causal graph and employ graph search techniques like A* \cite{astar} to find plans from some initial node in the graph to a desired goal node \cite{ff, fd, elephants}. Although these approaches thrive for purely symbolic domains, they do not naturally allow for the incorporation of numerical constraints and objective functions \cite{ivankovic2014optimal}. The formulation of AI Planning problems as Integer Programs has been explored a few times in the planning and scheduling literature \cite{vossen1999use, van2005optiplan}. 

In this work, we build up on Vossen et. al's \cite{vossen1999use} Integer Programming formulation by encoding the AI Planning problem as a MIP and solving it using off-the-shelf MIP solvers. This MIP encoding is convenient because it naturally allows for the incorporation of numerical constraints and objective functions into the planning problem. This ability is essential because real world problems often involve numerical constraints and objectives.  Metric-FF \cite{hoffmann2003metric} is a symbolic planning system that can account for numerical state variables. However, unlike our approach, Metric-FF is incapable of optimizing for convex objective functions, continuous action dynamics and continuous task constraints.

\subsection{Integrated Task and Motion Planning}
The class of approaches that interleave symbolic AI planning and continuous planning is called Task and Motion Planning \cite{shycobra, garrettfactored, lgp, srivastava}. Among these, approaches like Garrett et. al. \cite{garrettfactored} and Srivastava et. al. \cite{srivastava} devise symbols to describe continuous constraints for actions. These symbols are used as action preconditions in the symbolic AI planning process and are evaluated on demand. In addition to the chore of having to devise symbolic abstractions for every continuous constraint, these approaches are hampered by their requirement of symbolic goal descriptions. They are incapable of planning for continuous goal descriptions that can only be evaluated by an objective function. By formulating the planning problem as a Mixed Integer Convex Program, our proposed approach is able to both reason on the symbolic level using integer-valued variables and integer inequalities and optimize for continuous objective functions using the real-valued variables, and convex equations and inequalities.

\subsection{Combined symbolic and continuous planning as Mathematical Programs}
Works like Toussaint \cite{lgp} and Li and Williams \cite{kongming} have sought to solve the combined symbolic and continuous planning problem by formulating them as Mathematical Programs. Toussaint \cite{lgp} uses an iterative 3-level Nonlinear Constrained Optimization to optimize for continuous robot configurations over discrete action sequences it acquires from running Monte Carlo Tree Search. Our approach differs from Toussaint \cite{lgp} in that, we formulate the entire planning problem as a single Mixed Integer Program, solving for both the discrete action sequences and the continuous robot configurations in a single run.  Li and Williams \cite{kongming} employ hybrid flow graphs to represent the entire plan space and formulate the planning problem as a Mixed Logic Nonlinear Program in planning actions for autonomous underwater vehicles in ocean exploration tasks. Our representation of continuous actions as funnels and our formulation of Hybrid Funnel Graphs to represent the plan space are inspired by Li and Williams's work. 
\section{Problem Formulation}
In this work, we tackle the problem of optimal task planning under numerical constraints. Our goal is to generate a grounded plan made up of a logically consistent sequence of actions, with each action associated with its corresponding optimal continuous parameters. We will use the task of package rearrangement within a warehouse environment (The Warehouseman's Problem \cite{warehouse}) by a mobile manipulator robot as a running example for the remainder of this paper.

The inputs to our approach are:
\begin{itemize}
    \item A set of initial symbolic propositions, $\mathcal{I}$, that describe the initial symbolic state of the world. For example, the set
    \begin{equation*} 
    \mathcal{I} = \{\footnotesize{\texttt{(hand-empty),\\
                    (not (packed boxA))}} \}
    \end{equation*}
    represents a world where the robot is not holding any package and \texttt{boxA} has not been packed.
    
    \item A set of initial continuous variable values, $\mathcal{X}^I _r$ and $\mathcal{X}^I _b$, where $\mathcal{X}^I _r$ represents the robot's initial configuration in SE(2) space and $\mathcal{X}^I _b$ represents the configurations of the packages, also in SE(2) space.
    
    \item A set of action primitives, $\mathcal{A}$, that can be executed by a robot. An action primitive is comprised of:
    \begin{itemize}
        \item Symbolic preconditions: A conjunction of symbols whose truth-value must be true in order for the action to be executable.
        \item Continuous preconditions: Continuous constraints on the continuous variables ($\mathcal{X} _r$ and $\mathcal{X} _b$) that must be satisfied in order for the action to be executable.
        \item Action Dynamics: A dynamics function that computes the state of the continuous variables ($\mathcal{X} _r$ and $\mathcal{X} _b$) after the action is executed.
        \item Symbolic effects: A conjunction of symbols that represent the state of the world after the action is executed.
        \item Continuous effects: The numerical values of continuous variables ($\mathcal{X} _r$ and $\mathcal{X} _b$) after the action is executed.
    \end{itemize}
    \item A set of task-specific numerical constraints $\mathcal{H}$ on the continuous variables.
    \item A set of goal symbolic propositions, $\mathcal{G}$, that must hold true at the end of the plan execution. For example, the symbolic propositions
    \begin{equation*} 
    \mathcal{G} = \{\footnotesize{\texttt{(hand-empty),\\
                     (packed boxA)}} \}
    \end{equation*}
    would represent a world where the robot is not holding any package and \texttt{boxA} is packed.
    \item A set of goal continuous variable values, $\mathcal{X}^G _r$ and $\mathcal{X}^G _b$.
    \item An objective function $J(\mathcal{X} _r, \mathcal{X} _b)$ to be optimized.
\end{itemize}

The output of our approach is a grounded plan $\pi^*$ made up of a sequence of logically consistent actions $\{a_1(\mathcal{X}^{1*} _r, \mathcal{X}^{1*} _b), a_2(\mathcal{X}^{2*} _r, \mathcal{X}^{2*} _b), \dots , a_N (\mathcal{X}^{N*} _r, \mathcal{X}^{N*} _b) \}$, with each action associated with its corresponding optimal continuous parameter values.
\begin{figure*}
    \centering
    \includegraphics[width=0.8\textwidth]{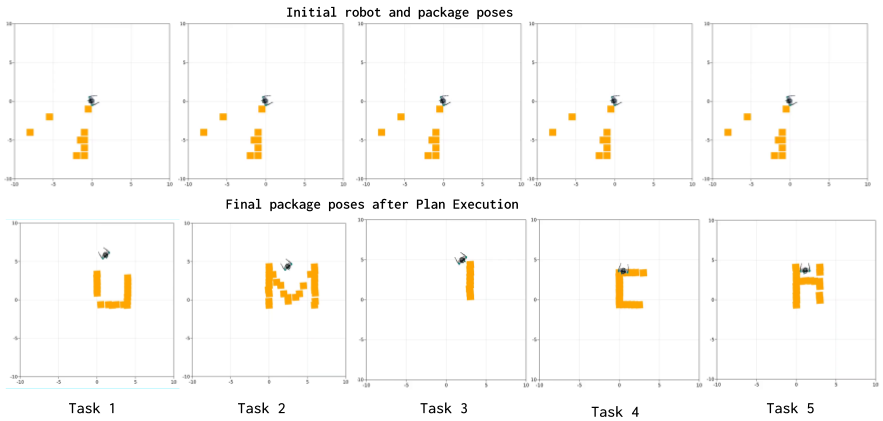}
    \caption{Qualitative results from the execution of plans generated by solving the package rearrangement problems in Tasks 1-5 with GTPMIP}
    \label{fig:warehouse}
\end{figure*}
\section{Methodology}
In this section, we describe each component of our approach, as illustrated in Figure \ref{fig:overview}. We name our approach \textit{Grounded Task Planning as Mixed Integer Programming} (GTPMIP). As stated in the previous section, GTPMIP takes as input a description of the optimal constrained task planning problem including descriptions of action primitives the robot is capable of executing. GTPMIP then builds a Hybrid Funnel Graph from this description to represent the entire plan space. Finally, it encodes the Hybrid Funnel Graph and the planning problem as an MICP, which it then solves using an off-the-shelf MIP solver. Each of these components are described in the following subsections.

\subsection{Hybrid PDDL Description}
Hybrid PDDL Description (HPD) is an extension of PDDL that allows for the specification of task planning problems with numerical action and task constraints, numerical initial values of continuous task variables, numerical objective functions, numerical action dynamics functions,  numerical  preconditions  and  numerical effects. Similar to PDDL, an HPD description of a task planning problem is made up of two files; the \texttt{domain.hpd} file and the \texttt{problem.hpd} file. 

The \texttt{domain.hpd} file describes the action primitives that the robot can execute. An action primitive has fields
\begin{itemize}
    \item \texttt{:action} to specify the name of the action primitive
    \item \texttt{:parameters} to specify the symbolic and continuous parameters the action takes.
    \item \texttt{:precondition} to specify a conjunction of symbols whose truth-values must be true in order for the action to be executable.
    \item \texttt{:continuous\_precondition} to specify continuous constraints on the continuous variables that must be satisfied in order for the action to be executable.
    \item \texttt{:dynamics} to specify dynamics functions that compute the state of the continuous variables after the action is executed.
    \item \texttt{:continuous\_effect} to specify the numerical values of continuous variables after the action is executed.
    \item \texttt{:effect} to specify a conjunction of symbols that represent the state of the world after the action  is executed.
\end{itemize}

The \texttt{problem.hpd} file describes the initial symbolic and continuous states as well as the goal symbolic and continuous states of the task. It also describes the task-specific constraints and the objective function to be optimized. 

PDDL+, which is also an extension to PDDL, allows for the specification of numerical action effects. However, unlike HPD,  PDDL+ is incapable of specifying numerical task constraints, objective functions and action dynamics functions.


\subsection{Funnels and Hybrid Funnel Graphs}
\begin{figure}
    \centering
    \includegraphics[width=0.45\textwidth]{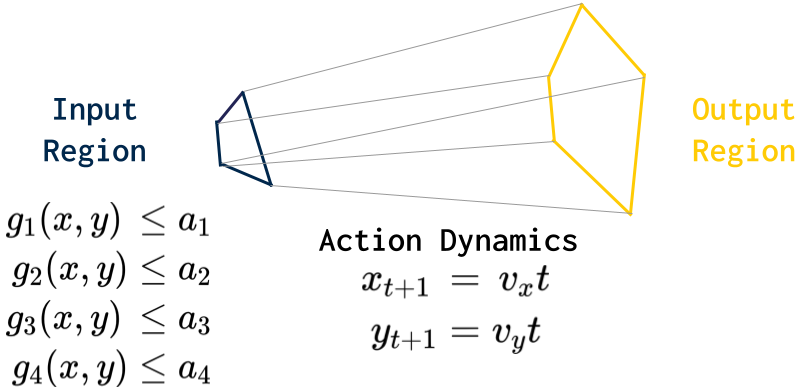}
    \caption{A funnel representation for the \texttt{move} action. The input region is formed by the intersection of continuous precondition inequality  constraints ($g_1 ~\text{to}~ g_4)$ on the robot position. The action dynamics compute the next robot position, $x_{t+1}, y_{t+1}$ after the action is applied to the poses in the input region. The output region represents the space of resulting poses.}
    \label{fig:funnel}
\end{figure}

We represent action primitives as \textit{funnels}. A funnel is made up of three components; an input region, a dynamics function and an output region. The input region is the region of intersection of all the continuous constraints that need to be satisfied before the action can be executed (the continuous preconditions). The dynamics function computes the state of the continuous variables after the action is executed. We apply  the dynamics function on the peripheries of the input region to result in a new region which we call the output region. The geometric representation of this abstraction takes the shape of a funnel as shown in Figure \ref{fig:funnel}; hence its name. The representation of action primitives as funnels helps in determining which action primitives are \textit{applicable} given the state of the robot. If the values of the continuous variables of the current state  intersects with the input region of a funnel and the symbolic preconditions of the corresponding action hold true for the symbolic propositions of current state, then the action is \textit{applicable}. The output region of the funnel also determines the continuous state after the corresponding action is executed. In addition to action funnels,  \textit{No-op} funnels are identity operations which represent actions that make no changes to the symbolic state of the world and whose set of symbolic preconditions are equal to their set of symbolic effects.

Given this representation of actions as funnels, we build up the Hybrid Funnel Graph by alternating between state levels and action levels. A state level is a set of all possible states (both symbolic and continuous) at a specific time instance. An action level is a set of all \textit{applicable} funnels at a specific time instance. The first state level is a set of all the symbolic propositions $\mathcal{I}$ and continuous variables $\mathcal{X}^I _b$ where $\mathcal{X}^I _r$ that make up the initial state. The continuous variables could take the form of either singular continuous values or intervals of continuous values that represent regions in the continuous space (SE(2) space in our package rearrangement problem formulation). We then compute all funnels that are applicable given the symbolic and continuous state variables in the first state level. These funnels constitute the first action level. As noted in the previous paragraph, a funnel is applicable to a state level if the continuous state variables in the state level intersect with the input region of the funnel and the symbolic preconditions of the action corresponding to the funnel hold true for the symbolic propositions of the state level. We also include to the first action level, \textit{No-op} funnels for each symbolic proposition in the state level.  The second state level is then computed as the set of all symbolic effects of actions and output regions of their corresponding funnels in the first action level. These output regions are computed by applying the funnel's dynamics function to the region of intersection of the continuous variables of the first state level and the funnel's input region. Likewise, the second action level is computed in the same manner as the first action level. 

After the computation of each state level, we check if the goal symbolic  propositions $\mathcal{G}$ hold true in the state level and if the goal continuous variable values $\mathcal{X}^G _b$ where $\mathcal{X}^G _r$ intersect the continuous variable regions in the state level. If both of these conditions are true, we have a valid Hybrid Funnel Graph for the task and terminate the graph building process. If not, we keep building the graph by adding additional state and action levels. Each action level represents a single time step in the resulting plan. Hence the total number of action levels, $T$, represents the total period of the entire resulting plan. Note that, since the Hybrid Funnel Graph starts with the initial state level and ends with the terminal state level, the total number of state levels is greater than the total number of action levels, $T$, by 1. This process is similar to the process of building planning graphs in GraphPlan \cite{graphplan} except that planning graphs in GraphPlan are made up of only symbolic propositions and symbolic actions. Hybrid Funnel Graphs are made up of both symbolic propositions and continuous variables, hence its name.

Unlike with GraphPlan where the graph building process is guaranteed to terminate if the planning problem is valid, our approach to building Hybrid Funnel Graphs is not guaranteed to terminate due to our inclusion of continuous variables. However in this work we observe GTPMIP successfully terminate graph building in every planning problem it is applied to.


\subsection{Encoding as a Mixed Integer Convex Program}
Before we encode the Hybrid Funnel Graph and the planning problem as an MICP, we first define a set of useful variables. 
\begin{itemize}
\item Let $T$ represent the total number of action levels in our Hybrid Funnel Graph, which is also the total planning period.
    \item Let $\mathcal{F}$ represent the set of all instantiated symbolic propositions in our planning domain. Hence $\mathcal{I} \subseteq \mathcal{F}$ and $\mathcal{G} \subseteq \mathcal{F}$
    \item Let $\mathcal{A}$ represent the set of all instantiated actions in our planning domain.
    \item Let $\text{pre}_f$ represent the set of all actions that have symbolic proposition $f$ as a symbolic precondition.
    \item Let $\text{add}_f$ represent the set of all actions whose symbolic effects affirm symbolic proposition $f$. (the truth-value of $f$ in the action's symbolic effect is \textbf{True}).
    \item Let $\text{del}_f$ represent the set of all actions whose symbolic effects negate symbolic proposition $f$. (the truth-value of $f$ in the action's symbolic effect is \textbf{False})
\end{itemize}

Next, we define integer variables. For all $f \in \mathcal{F}$ and $t \in 1 \dots T$,
\begin{equation*}
    p_{f, t}=
    \begin{cases}
      1, & \text{if proposition } f \text{ holds true at time } t \\
      0, & \text{otherwise}
    \end{cases}
\end{equation*}

\begin{equation*}
    q_{a, t}=
    \begin{cases}
      1, & \text{if action } a \text{ is taken at time } t \\
      0, & \text{otherwise}
    \end{cases}
\end{equation*}

Given these variable definitions, we now build the constraints into our MICP.

The first set of constraints to be added are the initial and terminal constraints. 

The initial constraint,
\begin{equation}
    p_{f, 1}=
    \begin{cases}
      1, & \forall f \in \mathcal{I} \\
      0, & \forall f \notin \mathcal{I}
    \end{cases}
\end{equation}
ensures that all initial symbolic propositions hold true in the first state level.

The terminal constraint, 
\begin{equation}
    p_{f, T+1} = 1, ~~\forall f \in \mathcal{G} 
\end{equation}
ensures that all goal symbolic propositions hold true in the last state level.

The next set of constraints are the precondition constraints
\begin{equation}
    q_{a, t} \leq p_{f, t}, ~~\forall a \in \text{pre}_f, \forall t \in 1 \dots T, f \in \mathcal{F}
\end{equation}
These inequality constraints encode the implication constraint that if action $a$, which has symbolic proposition $f$ as its precondition, is taken in action level $t$, then $f$ should also hold true in state level $t$. This constraint is called the precondition constraint because it ensures that all preconditions of an action hold true before the action can be taken.

The next set of constraints are the effect constraints
\begin{equation}
    p_{f, t+1} \leq \sum _{a \in \text{add}_f} q_{a, t}, ~~\forall t \in 1 \dots T, f \in \mathcal{F}
\end{equation}
These inequality constraints encode the implication constraint that if symbolic proposition $f$ holds true in state level $t+1$, then at least one action $a$ which has $f$ as a positive effect should be taken in action level $t$.

The next set of constraints are the mutual exclusion constraints
\begin{equation}
    q_{a, t} + q_{a', t} \leq 1
\end{equation}
for all $t \in 1 \dots T$ and all $a, a'$ for which there exists an $f \in \mathcal{F}$ such that $a \in \text{del}_f$ and $a' \in \text{pre}_f \cup \text{add}_f$.\\
These inequality constraints ensure that two actions $a$ and $a'$ that cancel each other are not both taken in action level $t$.

The next set of constraints are the task-specific numerical constraints
\begin{equation}
    q_{a, t} \leq h(\mathcal{X}^t _r, \mathcal{X}^t _b), ~~\forall h \in \mathcal{H}, t \in 1 \dots T
\end{equation}
that ensure that if action $a$ is taken in action level $t$, the continuous variable parameters of $a$ satisfy all the task-specific numerical constraints $\mathcal{H}$.

The final set of constraints are the initial and terminal constraints
\begin{equation}
\begin{split}
    \mathcal{X}^1 _r = \mathcal{X}^I _r, ~~
    \mathcal{X}^1 _b = \mathcal{X}^I _b
\end{split}
\end{equation}
and 
\begin{equation}
\begin{split}
    \mathcal{X}^{T+1} _r = \mathcal{X}^G _r, ~~
    \mathcal{X}^{T+1} _b = \mathcal{X}^G _b
\end{split}
\end{equation}
that ensure that the values of continuous variables at the first and final levels are equal to the problem-specified initial and goal continuous variable values respectively.

The objective function to be optimized 
\begin{equation}
J(\mathcal{X}^t _r, \mathcal{X}^t _b) ~~\forall t \in 1\dots T
\end{equation}
is a convex function on all continuous variables for the entire planning period. For a warehouseman's problem, a suitable objective function would be to minimize the total Euclidean distance the robot travels while rearranging the packages.

Putting together Equations 1 - 9, our entire MICP can now be summarized as 

\begin{equation}
\begin{split}
    \min_{\mathcal{X}^t_r, \mathcal{X}^t_b, q_{a, t}} J(\mathcal{X}^t _r, \mathcal{X}^t _b), ~~ \forall t \in 1\dots T\\
    \text{subject to \quad\quad\quad\quad\quad\quad\quad\quad} \\  
    p_{f, 1}=
    \begin{cases}
      1, & \forall f \in \mathcal{I} \\
      0, & \forall f \notin \mathcal{I}
    \end{cases}\\
    p_{f, T+1} = 1, ~~\forall f \in \mathcal{G} \\
    q_{a, t} \leq p_{f, t}, ~~\forall a \in \text{pre}_f, ~~\forall t \in 1 \dots T, f \in \mathcal{F}\\
    p_{f, t+1} \leq \sum _{a \in \text{add}_f} q_{a, t}, ~~\forall t \in 1 \dots T, f \in \mathcal{F}\\
    q_{a, t} + q_{a', t} \leq 1\\
    q_{a, t} \leq h(\mathcal{X}^t _r, \mathcal{X}^t _b), ~~\forall h \in \mathcal{H}, t \in 1 \dots T\\
    \mathcal{X}^1 _r = \mathcal{X}^I _r, ~~
    \mathcal{X}^1 _b = \mathcal{X}^I _b\\
    \mathcal{X}^{T+1} _r = \mathcal{X}^G _r, ~~
    \mathcal{X}^{T+1} _b = \mathcal{X}^G _b\\
    p \in \{0, 1\}, q \in \{0, 1\}, \mathcal{X}\in \text{SE(2)}\\ 
\end{split}
\end{equation}

We solve this MICP using an off-the-shelf MIP solver which returns the grounded plan $\pi^*$ made up of a sequence of logically consistent actions $\{a_1(\mathcal{X}^{1*} _r, \mathcal{X}^{1*} _b), a_2(\mathcal{X}^{2*} _r, \mathcal{X}^{2*} _b), \dots , a_T (\mathcal{X}^{T*} _r, \mathcal{X}^{T*} _b) \}$, with each action associated with its corresponding optimal continuous parameter values.

\section{Implementation}
\subsection{Open-source software implementations}
We provide open-source software implementations for each of the components of our approach. We provide
\begin{itemize}
    \item \small{\texttt{\textbf{HPD.jl}}} as a software package for reading and parsing HPD files. It also contains example HPD files for the warehouse rearrangement problem. url: \href{https://github.com/adubredu/HPD.jl}{https://github.com/adubredu/HPD.jl}
    \item \small{\texttt{\textbf{HybridFunnelGraphs.jl}}} as a software package for building complete Hybrid Funnel Graphs when given HPD files of an optimal constrained Task Planning Problem. url: \href{https://github.com/adubredu/HybridFunnelGraphs.jl}{https://github.com/adubredu/HybridFunnelGraphs.jl}
    \item \small{\texttt{\textbf{gtpmip.jl}}} as a software package for encoding Hybrid Funnel Graphs and HPD files of an optimal constrained task planning problem as an MIP and solving the MIP to output the optimal plan. url: \href{https://github.com/adubredu/gtpmip.jl}{https://github.com/adubredu/gtpmip.jl}
    \item \small{\texttt{\textbf{westbrick.jl}}} as a simulator for a 2D version of the Warehouse rearrangement problem. url: \href{https://github.com/adubredu/westbrick.jl}{https://github.com/adubredu/westbrick.jl}
\end{itemize} 

\subsection{Solving the Mixed Integer Convex Program}
Throughout our experiments, we use the Gurobi Optimization software \cite{gurobi} to solve all MICPs. We use Gurobi because it was the fastest MIP solver amongst all solvers considered.
\section{Experiments}
We evaluate the capabilities of GTPMIP on a series of experiments both in simulation and on a physical robot.

\subsection{Pure Symbolic Task Planning evaluation}
First, we compare the symbolic task planning capabilities of GTPMIP to the state-of-the-art symbolic planning approaches Fast-Downward \cite{fd},  Pyperplan \cite{pyperplan} and Forward Search with A* \cite{forward}. We compare the planning times of these approaches on purely symbolic block stacking tasks with increasing problem size, with Problem 1 having 4 blocks and Problem 5 having 9 blocks. Table \ref{table:symbexp} shows the average planning times of each approach. 

\begin{table*}[]
    \centering
    \begin{tabular}{|c||c|c|c|c|c|}
    \hline
         \textbf{Algorithm} & \textbf{Problem 1} & \textbf{Problem 2} & \textbf{Problem 3} & \textbf{Problem 4} & \textbf{Problem 5}  \\
         \hline \hline 
         Fast Downward \cite{fd}& $0.102 \pm 0.0003$ & $0.104 \pm 0.002$ & $0.214 \pm 0.001$ & $0.218 \pm 0.002$ & $0.216 \pm 0.008$ \\
        \hline         
         Pyperplan \cite{pyperplan} & $0.167 \pm 0.008$ & $0.169 \pm 0.012$ & $0.169 \pm 0.014$ & $0.183 \pm 0.012$ & $0.190 \pm 0.007$\\
         \hline
         Forward Search \cite{forward}& $0.0006 \pm 0.004$ & $0.0011 \pm 0.001$ & $0.0021 \pm 0.0014$ & $0.004 \pm 0.002$ & $0.008 \pm 0.004$\\
         \hline
         \textbf{GTPMIP (ours)} & \textbf{$0.039 \pm 0.011$} &\textbf{ $0.033 \pm 0.0004$} & \textbf{$0.062 \pm 0.009$} & \textbf{$0.172 \pm 0.011$} & \textbf{$0.719 \pm 0.014$}\\
         \hline
    \end{tabular}
    \caption{Comparison of planning times (in seconds) of GTPMIP with those of state-of-the-art symbolic planners on purely symbolic block stacking problems with increasing number of blocks. Problem 1 and Problem 2 have 4 blocks, Problem 3 has 5 blocks, Problem 4 has 6 blocks and Problem 5 has 9 blocks. }
    \label{table:symbexp}
\end{table*}

As can be seen from results in Table \ref{table:symbexp}, GTPMIP is slightly faster than Fast Downward and Pyperplan on the smaller Problems 1 - 4. GTPMIP however gets much slower than the other symbolic planning algorithms as the size of the problem increases in Problem 5. This significant reduction in planning speed can be attributed to the increase in number of variables and constraints in the resulting MIP that GTPMIP solves. However, the unique capabilities of GTPMIP that are lacking in the other symbolic planning approaches are its ability to account for numerical constraints and optimize for numerical objective functions. This is demonstrated in the next experiment.

\subsection{Warehouse package rearrangement Problem}
Next, we evaluate GTPMIP on a series of 5 tasks to evaluate its ability to perform optimal task planning under numerical constraints. Each task is setup with a virtual robot in a 2D warehouse simulator. For each Warehouse package rearrangement problem, the goal is to plan for the optimal action sequence with optimal continuous parameters that rearrange the packages by satisfying a specific set of linear geometric constraints on package placements in SE(2) space. The set of constraints for the five tasks are listed in Table \ref{table:constraints}. 

We evaluate the time to build the Hybrid Funnel Graph as well as the time to solve the resulting MICP for each of these tasks. Quantitative experimental results for each task are presented in Table \ref{table:warehouse}, with the corresponding qualitative results shown in Figure \ref{fig:warehouse}. The experiments were run in \texttt{westbrick.jl}, a 2D package rearrangement simulator we developed.  

Videos of the robot executing the plans generated for each task can be found on the project's webpage at this url: \href{https://adubredu.github.io/gtpmip}{https://adubredu.github.io/gtpmip}

\begin{table}[t]
\centering
    \begin{tabular}{|l||c|}
    \hline
    \textbf{Task} & \textbf{Constraints} \\
    \hline\hline
     \multirow{3}{4em}{Task 1}    &  $x = 0.0 ~~\cap~~ 0.0 \leq y \leq 4.0$ \\
                                  &   $0.0 \leq x \leq 4.0 ~~\cap~~ y = 0.0$ \\
                                  &  $x = 4.0 ~~\cap~~ 0.0 \leq y \leq 4.0$ \\
    \hline
    \multirow{4}{6em}{Task 2}     & $x = 0.0 ~~\cap~~ y = 5.0$ \\
                                  & $x = 6.0 ~~\cap~~ y = 5.0$ \\
                                  & $4.5 \leq y + 1.6x \leq 5.0 ~~\cap~~ \dots$\\
                                  & $0.0 \leq x \leq 3.0 ~~\cap~~ 0.0 \leq y \leq 4.0$ \\
                                  & $-5.0 \leq y - 1.6x \leq -4.5 ~~\cap~~ \dots$\\
                                  & $3.0 \leq x \leq 6.0 ~~\cap~~ 0.0 \leq x \leq 4.0$\\
    \hline
    \multirow{1}{4em}{Task 3}     & $x = 3.0 ~~\cap~~ 0.0 \leq y \leq 6.0$\\
    \hline 
    \multirow{3}{4em}{Task 4}     & $x = 0 ~~\cap~~ 0.0 \leq y \leq 4.0$\\
                                  & $0.0 \leq x \leq 3.0 ~~\cap~~ y = 4.0$\\
                                  & $0.0 \leq 3.0 ~~\cap~~ y = 0.0$\\
    \hline 
    \multirow{3}{4em}{Task 5}     & $x = 0.0 ~~\cap~~ 0.0 \leq y \leq 5.0$\\
                                  & $0.0 \leq x \leq 3.0 ~~\cap~~ y = 4.0$ \\
                                  & $x = 3.0 ~~\cap~~ 0.0 \leq y \leq 5.0$ \\
    \hline
    \end{tabular}
    \caption{The set of geometric constraints on package placements for each of the Warehouse package rearrangement tasks.}
    \label{table:constraints}
\end{table}

\begin{table}[]
    \centering
    \begin{tabular}{|c||c|c|}
        \hline
       \textbf{Task} & \textbf{HFG Building Time(s)}  &  \textbf{MICP Solving Time(s)}\\
       \hline \hline
       Task 1 & $21.12 \pm  1.230$ &  $0.26 \pm 0.100$ \\
       \hline
       Task 2 & $87.41 \pm 10.160$ & $0.60 \pm 0.084$ \\
        \hline
       Task 3 & $5.20 \pm 0.140$ & $0.12 \pm 0.004$ \\
        \hline
       Task 4 & $29.92 \pm 0.450$ & $0.31 \pm 0.044$\\
       \hline
       Task 5 & $32.06 \pm 0.330$ & $0.30 \pm 0.046$\\
       \hline
    \end{tabular}
    \caption{Times (in seconds) for building the Hybrid Funnel Graphs(HFG) and for solving the resulting Mixed Integer Convex Program for each of the tasks.}
    \label{table:warehouse}
\end{table}


\subsection{Mobile Manipulation tasks with Physical Robot}
Finally, we employ GTPMIP in planning for optimal constrained tasks in the real world. We use the Digit \cite{digit} humanoid robot platform to execute output plans. We focus on 2 main tasks; the shelf-stocking task (pictured in Figure \ref{fig:teaser}) and the table serving task.  

The shelf-stocking task requires that the robot stock a shelf with a predefined set of grocery items at specific positions on the shelf. The table serving task requires that the robot collects a specified set of grocery items from a shelf and distributes them to a dinner table in predefined desired configurations.  

The kinematic constraints we consider in these tasks are the robot stance pose constraint and the grasp angle constraint. The robot stance pose constraint constrains the robot's standing pose to a desired region in SE(2) space from which the object to be grasped is kinematically reachable by the robot. The grasp angle constraint ensures that the angle of approach of the robot's grippers results in a stable grasp. 

Video demonstrations of the robot performing all these tasks can be found on the project's website at this url: \href{https://adubredu.github.io/gtpmip}{https://adubredu.github.io/gtpmip}

\section{Discussion}
The primary assumption we made in this work is that the entire environment was fully-observable; that the robot had absolute knowledge about the poses of all objects of interest, the best constraints to satisfy and the right set of action primitives. However, robots operating in most interesting real world settings do not often have access to these capabilities. An exciting avenue for future work would be to ease the expense of predefining constraints. Could we learn to derive numerical constraints from natural language or from user demonstrations?
Another potential avenue for future focus would be to enable robots to autonomously learn the right set of action primitives needed to complete a given task.  
\section{Conclusion}
We tackled the problem of optimal constrained task planning by proposing an approach that encoded the entire task planning problem as a single MICP and solved it using an off-the-shelf MIP solver. We evaluated our approach on a set of optimal constrained task planning problems and demonstrated its ability to generate optimal plans including on a physical robot platform under kinematic constraints.
\bibliographystyle{plain}
\bibliography{references}

\end{document}